\documentclass{article}
\usepackage[preprint]{neurips_2024} % put 'preprint' here, not on \documentclass

\usepackage[utf8]{inputenc}
\usepackage[T1]{fontenc}
\usepackage{setspace}
\usepackage{graphicx}
\usepackage{float}
\usepackage{booktabs}
\usepackage{subcaption}  % modern subfigures
\usepackage{longtable}
\usepackage{array}
\usepackage{caption}
\usepackage{booktabs}

\title{Towards Synthesizing Normative Data for Cognitive Assessments Using Generative Multimodal Large Language Models}
\author{%
\textbf{Victoria Yan}$^{1}$\thanks{This work was conducted during the internship at the Center for Data Science, Nell Hodgson Woodruff School of Nursing, Emory University} \quad
\textbf{Honor Chotkowski}$^2$ \quad
\textbf{Fengran Wang}$^3$ \quad
\textbf{Xinhui Li}$^4$ \quad
\textbf{Carl Yang}$^3$ \quad\\
\textbf{Jiaying Lu}$^2$ \quad
\textbf{Runze Yan}$^2$ \quad
\textbf{Xiao Hu}$^2$ \quad
\textbf{Alex Fedorov}$^2$\thanks{Corresponding author: \texttt{avfedor@emory.edu}} \\
$^1$The Westminster Schools \\
$^2$Center for Data Science, Nell Hodgson Woodruff School of Nursing, Emory University \\
$^3$Department of Computer Science, Emory University \\
$^4$School of Electrical and Computer Engineering, Georgia Institute of Technology\\
}
\date{}  % Optional: leave empty for no date

\begin{document}

\maketitle

%\noindent\textbf{Note:} Original Research Articles are limited to 2700 words, include an abstract, a maximum of five tables or figures (total), and up to 40 references. All text, references, figure legends, and tables should be included in this file. Figures may also be embedded for the first submission.

% ===== Structured Abstract =====
\begin{abstract}
    \textbf{Background:} Cognitive assessments require normative data as essential benchmarks for evaluating individual performance. Hence, developing new cognitive tests based on novel image stimuli is challenging due to the lack of readily available normative data. Traditional data collection methods are costly, time-consuming, and infrequently updated, limiting their practical utility. Recent advancements in generative multimodal large language models (MLLMs) offer a new approach to generate synthetic normative data from existing cognitive test images.

    \textbf{Methods:} We investigated the feasibility of using MLLMs, specifically GPT-4o and GPT-4o-mini, to synthesize normative textual responses for established image-based cognitive assessments, such as the "Cookie Theft" picture description task. Two distinct prompting strategies—naive prompts with basic instructions and advanced prompts enriched with contextual guidance—were evaluated. Responses were analyzed using embeddings to assess their capacity to distinguish diagnostic groups (controls, mild cognitive impairment, Alzheimer's disease) and demographic variations. Performance metrics included BLEU, ROUGE, BERTScore, and an LLM-as-a-judge evaluation.

    \textbf{Results:} Advanced prompting strategies produced synthetic responses that more effectively distinguished between diagnostic groups and captured demographic diversity compared to naive prompts. Superior models generated responses exhibiting higher realism and diversity. BERTScore emerged as the most reliable metric for contextual similarity assessment, while BLEU was less effective for evaluating creative outputs. The LLM-as-a-judge approach provided promising preliminary validation results.

    \textbf{Conclusions:} Our study demonstrates that generative multimodal LLMs, guided by refined prompting methods, can feasibly generate robust synthetic normative data for existing cognitive tests, thereby laying the groundwork for developing novel image-based cognitive assessments without the traditional limitations.
    \end{abstract}

% ===== Main Sections =====
\section{Background}

Normative data serve as the essential benchmarks for interpreting individual performance on cognitive assessments. Traditionally, such data have been collected through extensive manual efforts involving standardized testing administered to large, representative samples~\citep{Capitani1997, Mitrushina2005}. Conventional approaches for data collections often encounter significant challenges. These include high costs, lengthy timelines for data collection, and difficulties in maintaining the representativeness and relevance of norms amidst ongoing demographic and societal changes~\citep{delCachoTena2024}

Recent advances in generative modeling offer a promising alternative to traditional normative data collection. For instance, studies have demonstrated that generative frameworks can synthesize can generate synthetic data that accurately reflects the patterns in real clinical data~\citep{Chintapalli2024, Barr2025}. Such techniques have been applied to neuroimaging data, where synthetic samples produced by generative models closely approximate normative values, potentially reducing the effort needed to update traditional normative datasets.

Within the broader category of generative models, large language models (LLMs) specifically have advanced natural language processing (NLP) in medicine. State-of-the-art LLMs, including GPT-4, have shown strong capabilities in extracting structured information from unstructured clinical narratives and in processing multimodal data~\citep{OpenAI2023, Yin2024}. Given that many cognitive assessments, like the 'Cookie Theft' task, rely on interpreting visual stimuli, models capable of processing both images and text (multimodal LLMs) are particularly relevant. These models could potentially generate realistic textual responses directly from the image prompts used in such tests.Investigating whether these models can generate synthetic normative responses for cognitive assessments based on image stimuli and textual inputs motivates this study.

Nevertheless, the utilization of synthetic data generated by LLMs introduces additional challenges. Ethical concerns such as bias, error propagation, and reproducibility need to be thoroughly evaluated to ensure that the synthetic normative data are both valid and reliable~\citep{Smolyak2024, Susser2024}. Such issues underscore the necessity for robust validation protocols when applying generative models to domains as sensitive as cognitive assessment and clinical research.

In this work, we explore the feasibility of using generative multimodal LLMs to synthesize normative data for existing cognitive tests. We employ refined prompting strategies to guide the generation process and use a combination of quantitative metrics and an LLM-as-a-Judge evaluation approach to assess the realism, diversity, and diagnostic differentiation capabilities of the generated responses.

% -------------------------------------

\section{Methods}
\subsection{Study Design}

\begin{figure}[t]
    \centering
    \includegraphics[width=0.6\textwidth]{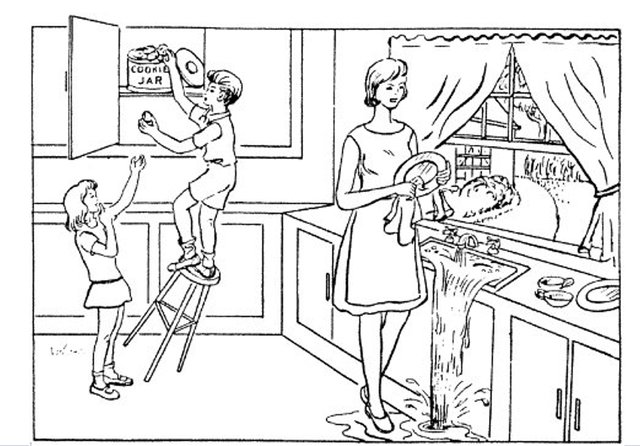}
    \caption{The "Cookie Theft" picture used in the narrative elicitation task.}
    \label{fig:cookie-theft}
\end{figure}

Our study was designed to evaluate the potential of generative multimodal large language models (MLLMs) to synthesize normative responses for image-based cognitive assessments. The process followed three key phases:

\begin{enumerate}
    \item \textbf{Data Conditioning:} We first selected demographic and cognitive metadata (e.g., age, gender, MMSE score, and cognitive diagnosis) from real participants in the Pitt Corpus~\citep{becker1994natural, lanzi2023dementiabank}. These data points were used to populate either naive or advanced prompting templates designed to simulate realistic patient profiles during a cognitive screening task.

    \item \textbf{Response Generation:} Using these prompts, we guided two MLLMs (GPT-4o and GPT-4o-mini) to generate simulated speech responses to the "Cookie Theft" image (Figure~\ref{fig:cookie-theft}). The responses were intended to reflect varying levels of cognitive ability, ranging from normal cognition to mild cognitive impairment (MCI) and Alzheimer's disease (AD).

    \item \textbf{Evaluation:} The generated responses were evaluated using standard NLP metrics (BLEU, ROUGE, BERTScore) and vector-based embedding techniques for clustering and semantic similarity. Additionally, we employed an \textit{LLM-as-a-Judge} method. This involved crafting an evaluation prompt that explicitly encoded annotation criteria we developed (e.g., coherence, relevance, complexity), enabling the LLM to serve as an automated evaluator of its own and peer model outputs.
\end{enumerate}

This design allowed us to assess both the feasibility and fidelity of synthetic normative data generation in the context of cognitive screening.

\subsection{Data Collection and Generation}

For this study, we used the Pitt Corpus~\citep{becker1994natural} from the DementiaBank database~\citep{lanzi2023dementiabank}, a widely cited open-access repository that supports the analysis of language decline in aging and dementia. The Pitt Corpus consists of audio recordings and transcriptions from clinical interviews collected as part of a longitudinal study on language and cognitive changes in older adults. It includes participants with Alzheimer's disease, mild cognitive impairment (MCI), and healthy controls. Full demographic and diagnostic details for our study subset are provided in Table X.

Our primary task involved the analysis and generation of narratives elicited from the well-established "Cookie Theft" picture description task, originally part of the Boston Diagnostic Aphasia Examination. Participants are shown a picture depicting a complex scene and asked to describe everything they see happening. This open-ended task allows for naturalistic speech samples that can reveal subtle markers of cognitive-linguistic decline. An example of the Cookie Theft image used in this task is shown in Figure~\ref{fig:cookie-theft}.

All data access and usage were conducted in compliance with the data sharing policies of the DementiaBank project. As the data were previously collected and de-identified, this study was determined to be exempt from further human subjects review.

\subsection{Generative Multimodal Models and Prompting Strategies}

We utilized OpenAI's generative multimodal large language models, specifically \textbf{GPT-4o} and its computationally efficient variant \textbf{GPT-4o-mini}, to simulate verbal responses in a dementia screening context. These models were tasked with generating plausible language output from individuals with varying cognitive statuses, based on minimal demographic and clinical metadata.

Two prompting strategies were employed to drive the models: a naive template and an advanced task-specific instructional prompt.

\textbf{Naive Prompting Strategy.} This method used a brief template with basic demographic and screening information. It served as a lightweight, general-purpose input to elicit spontaneous descriptive responses. The template is shown in Figure~\ref{fig:naive-prompt}.

\textbf{Advanced Prompting Strategy.} To better constrain and contextualize MLLM model behavior, we developed a comprehensive prompt that embeds the screening protocol, timing constraints, cognitive profiles, and task-specific instructions. This approach enabled more nuanced response generation and alignment with clinically observed language patterns in Alzheimer's disease and MCI. The advanced prompt is shown in Figure~\ref{fig:advanced-prompt}.

\begin{figure}[h]
    \centering
    \begin{minipage}{\textwidth}
    \small
    \begin{verbatim}
    def create_dementia_screening_prompt(age, gender, MMSE, category):
        age_text = f" {age}-year-old" if age else "of unknown age"
        gender_text = gender if gender else "of unspecified gender"
        mmse_text = f" with an MMSE score of {mmse}" if mmse else "with an unknown MMSE score"
        category_text = f" categorized as {category}" if category
        else "with an unspecified cognitive status"

        return (
            f "A clinician is screening for language-based cognitive impairment. "
            "The patient is shown an image and asked to describe everything happening in it. "
            f"A {age_text} {gender_text} {mmse_text}, {category_text}, is screened. "
            "What response might they produce?"
        )
    \end{verbatim}
    \caption{Naive dementia screening prompt.}
    \label{fig:naive-prompt}
    \end{minipage}
    \end{figure}

    \begin{figure}[h]
    \centering
    \begin{minipage}{\textwidth}
    \small
    \begin{verbatim}
    def generate_dementia_screening_prompt(age, gender, MMSE, category):
        prompt = f"""
        A clinician is conducting a screening for language-based cognitive
        impairment using an AI analysis tool. The patient is shown an image and asked:
        "Tell me what is happening in the following image."

        Task Details:
        The patient must begin speaking immediately in a spontaneous, conversational manner,
        A 90-second timer starts when the image is revealed. The transcript ends at the 90-second mark,
        The response must focus only on describing observable details in the image,
        Do not include phrases like "in the image" or commentary on artistic style,
        symbolism, or abstract interpretations,

        Patient Profile:
        Age: {age}
        Gender: {gender}
        MMSE Score: {mmse}
        Category: {category}

        Response Characteristics:
        Depending on the patient's cognitive condition,
        the response may exhibit one or more of the following characteristics:
        Alzheimer's Disease (Mild to Severe):
        Limited detail and reduced vocabulary,
        Anomia (word-finding difficulty),
        Disorganized or fragmented speech,
        Omission of key elements,
        Repetitive or circular speech patterns,
        Paraphasias (word substitutions or errors),
        Poor self-monitoring and off-topic speech,
        Agrammatism (grammatical errors),

        Mild Cognitive Impairment (MCI):
        Occasional word-finding difficulties,
        Subtle omissions of details,
        Mild disorganization in sequencing,

        No Impairment:
        Clear, coherent, and structured description,
        Comprehensive coverage of image details,

        Task: Simulate a possible patient response as if they were describing
        the image under the given conditions.
        The response should mimic natural, spontaneous speech, capturing any
        cognitive characteristics associated with the patient's condition.
        Do not generate any supporting explanation as to why large language models
        did that, return the response only.
        """
        return prompt
    \end{verbatim}
    \caption{Advanced dementia screening prompt with embedded clinical context and constraints.}
    \label{fig:advanced-prompt}
    \end{minipage}
    \end{figure}

\subsection{Evaluation}

\paragraph{BLEU (Bilingual Evaluation Understudy)}
BLEU measures the precision of n-gram overlaps between a candidate and reference text, incorporating a brevity penalty to discourage overly short outputs. It is widely used in machine translation but may not capture semantic adequacy or fluency~\citep{papineni2002bleu}.

\paragraph{ROUGE (Recall-Oriented Understudy for Gisting Evaluation)}
ROUGE evaluates text summaries by measuring the recall of n-gram overlaps, longest common subsequences, and skip-bigrams between candidate and reference summaries. It emphasizes recall, making it suitable for summarization tasks~\citep{lin2004rouge}.

\paragraph{BERTScore}
BERTScore computes similarity scores between candidate and reference texts using contextual embeddings from pre-trained models like BERT. It captures semantic similarity beyond exact word matches and correlates well with human judgments~\citep{zhang2019bertscore}.

\paragraph{Embedding Techniques}
Embedding techniques convert textual data into numerical vectors. Traditional methods like Word2Vec and GloVe produce static embeddings, while contextual embeddings such as Sentence-BERT~\citep{reimers2019sentencebert} provide dynamic representations based on surrounding text, enhancing performance in many NLP tasks. Dimensionality reduction techniques like t-SNE~\citep{van2008visualizing} are used to visualize these high-dimensional embeddings for qualitative assessment of clustering by diagnosis and demographics.

\paragraph{LLM-as-a-Judge Evaluation Method}
The LLM-as-a-Judge approach utilizes large language models to evaluate the outputs of other models by prompting them to score according to specific criteria (e.g., hallucinations, coherence, vocabulary, tone, and length). It offers scalable, consistent evaluations and shows a strong correlation with human judgments~\citep{zheng2023judging}.

\paragraph{Classification Performance}
We trained diagnostic classifiers on Sentence-BERT embeddings to distinguish between Control, Mild Cognitive Impairment (MCI), and Alzheimer's Disease (AD) groups. Models included elastic-net logistic regression and XGBoost, with hyperparameters optimized via stratified 5-fold cross-validation and Optuna trials. Performance was measured using one-versus-one macro-averaged ROC AUC on held-out test sets, reporting median AUC and interquartile ranges to capture variability across runs.

\paragraph{Word-Frequency Analysis}
To assess the lexical coverage of key scene elements, we performed word-frequency analyses on both synthetic and real response corpora. We tokenized each corpus and extracted frequencies for critical terms (e.g., \emph{cookie}, \emph{sink}, \emph{dish}), dynamic verbs (e.g., \emph{reach}, \emph{fall}), and interpersonal references (e.g., \emph{boy}, \emph{girl}). Results were visualized via word clouds and summarized in top-10 frequency tables to highlight differences in lexical emphasis across models and prompting strategies.

% -------------------------------------

\section{Results and Discussion}

\begin{table}
\caption{Median (IQR) metrics per clinical group. BERT F1 remains in the 0.80–0.86 range, indicating strong semantic alignment. ROUGE, a summarization-overlap metric, and Google-BLEU, a translation metric, both yield lower scores because the synthetic and reference answers are not expected to overlap. Comprehensive-prompt models (4o-comp, 4o-mini-comp) outperform the trivial-prompt model. 4o-comp also improves over 4o-mini-comp. Word counts are highest for 4o-mini-comp, but the distribution is further from the real and lowest for 4o-trivial. All metrics are lowest in the Alzheimer's-disease (AD) group.}
\centering
\small
\begin{tabular}{lllllll}
\toprule
Category & Model & ROUGE-1 & ROUGE-L & BERT F1 & Google BLEU & Num Words \\
\midrule
Control & 4o-comp & 0.28 (0.25–0.32) & 0.26 (0.23–0.29) & 0.85 (0.84–0.86) & 0.10 (0.08–0.12) & 102 (93–113) \\
 & 4o-mini-comp & 0.25 (0.21–0.29) & 0.24 (0.20–0.27) & 0.84 (0.83–0.85) & 0.08 (0.06–0.09) & 123 (111–136) \\
 & 4o-trivial & 0.26 (0.23–0.30) & 0.24 (0.20–0.27) & 0.84 (0.84–0.85) & 0.09 (0.08–0.11) & 104 (94–114) \\
 & Real Data & & & & & 92 (72–132) \\
\hline
MCI
 & 4o-comp & 0.27 (0.23–0.30) & 0.24 (0.21–0.29) & 0.85 (0.84–0.86) & 0.09 (0.07–0.10) & 103 (93.25–113) \\
 & 4o-mini-comp & 0.22 (0.19–0.25) & 0.21 (0.18–0.23) & 0.82 (0.82–0.83) & 0.07 (0.05–0.08) & 129 (119.25–136) \\
 & 4o-trivial & 0.26 (0.22–0.29) & 0.22 (0.19–0.25) & 0.83 (0.82–0.84) & 0.07 (0.06–0.08) & 76.50 (66–87) \\
  & Real Data & & & & & 91.50 (65–131) \\
\hline
AD
 & 4o-comp & 0.20 (0.16–0.23) & 0.19 (0.15–0.23) & 0.82 (0.81–0.83) & 0.06 (0.05–0.07) & 92 (80–104) \\
 & 4o-mini-comp & 0.21 (0.17–0.24) & 0.20 (0.16–0.23) & 0.82 (0.81–0.83) & 0.06 (0.04–0.07) & 127 (116.50–139) \\
 & 4o-trivial & 0.14 (0.10–0.18) & 0.13 (0.09–0.17) & 0.81 (0.80–0.82) & 0.05 (0.03–0.06) & 70 (55–93.50) \\
 & Real Data & & & & & 86 (58–116) \\
\hline
\bottomrule
\end{tabular}
\label{tab:metrics}
\end{table}

\subsection{Comparison of Synthetic Data Generation to the Real Data}

Synthetic clinical speech responses were generated using three distinct prompting strategies: a comprehensive prompt (4o-comp), a minimized comprehensive prompt (4o-mini-comp), and a trivial prompt (4o-trivial). These synthetic outputs were evaluated by comparing them to responses from real clinical samples matched by demographic characteristics, categorized into Control, Mild Cognitive Impairment (MCI), and Alzheimer's Disease (AD) groups. Four metrics were employed to assess synthetic responses: ROUGE-1 and ROUGE-L for lexical overlap common in summarization tasks, Google BLEU for translation-like lexical precision, and BERT F1 to evaluate semantic alignment through contextual embedding similarity. Results are presented as median values with interquartile ranges (IQR) to accurately reflect distributions and variability in Table~\ref{tab:metrics}.

Performance varied notably across the clinical groups, with metrics consistently highest in the Control group and progressively lower in MCI and Alzheimer's Disease (AD) groups. Semantic alignment, measured by BERT F1 scores, was highest for the Control group using the 4o-comp model (median = 0.85, IQR = 0.84–0.86) and lowest for the AD group (median = 0.82, IQR = 0.81–0.83). Lexical overlap metrics such as ROUGE-L showed a similar trend, decreasing from 0.26 (IQR = 0.23–0.29) in the Control group to 0.19 (IQR = 0.15–0.23) in the AD group for the 4o-comp model. Likewise, Google BLEU scores declined from 0.10 (IQR = 0.08–0.12) in Control subjects to 0.06 (IQR = 0.05–0.07) in the AD group. This pattern of decreasing performance likely reflects the increasing complexity and variability in responses characteristic of cognitive impairment groups, making accurate alignment with reference clinical responses more challenging.

\subsection{Effectiveness of Prompting Strategies}

The comprehensive prompting strategies (4o-comp and 4o-mini-comp) consistently outperformed the trivial prompting (4o-trivial) across clinical groups. In the MCI group, the median BERT F1 for 4o-comp was higher (0.85, IQR = 0.84–0.86) compared to 4o-trivial (0.83, IQR = 0.82–0.84). Similarly, the 4o-comp achieved higher ROUGE-1 scores (median = 0.20, IQR = 0.16–0.23) compared to trivial prompting (median = 0.14, IQR = 0.10–0.18) within the AD group. Additionally, the comprehensive models produced word counts more closely aligned with real clinical responses, for example, in the Control group, the 4o-comp had a median word count of 102 (IQR = 93–113), closely matching the median of real data at 92 (IQR = 72–132), whereas trivial prompting showed less realistic distributions (median = 104, IQR = 94–114). These results indicate that detailed, comprehensive prompts improve both semantic alignment and realism in terms of response length.

\subsection{Comparison Between OpenAI GPT 4o and 4o-mini Models}

Between the two comprehensive prompting models, the 4o-comp consistently outperformed the 4o-mini-comp, showing higher semantic and lexical alignment despite generating fewer words but closer to the real data. For instance, in the Control group, the median ROUGE-1 score for the 4o-comp model was 0.28 (IQR = 0.25–0.32), compared to 0.25 (IQR = 0.21–0.29) for the 4o-mini-comp. BERT F1 also favored the 4o-comp model (median = 0.85, IQR = 0.84–0.86) over the 4o-mini-comp (median = 0.84, IQR = 0.83–0.85). Although 4o-mini-comp produced higher word counts—123 words (IQR = 111–136) in the Control group compared to 102 words (IQR = 93–113) for 4o-comp—the increased word count did not correspond to improved semantic alignment or realism. Rather, the 4o-comp generated responses more closely matched both the semantic content and word-count distributions observed in real clinical samples, suggesting the advantage of utilizing a better model.

\subsection{Comparison to Real-World Responses}
\begin{figure}[htbp]
    \centering

    \begin{subfigure}[b]{0.3\linewidth}
        \includegraphics[width=\linewidth]{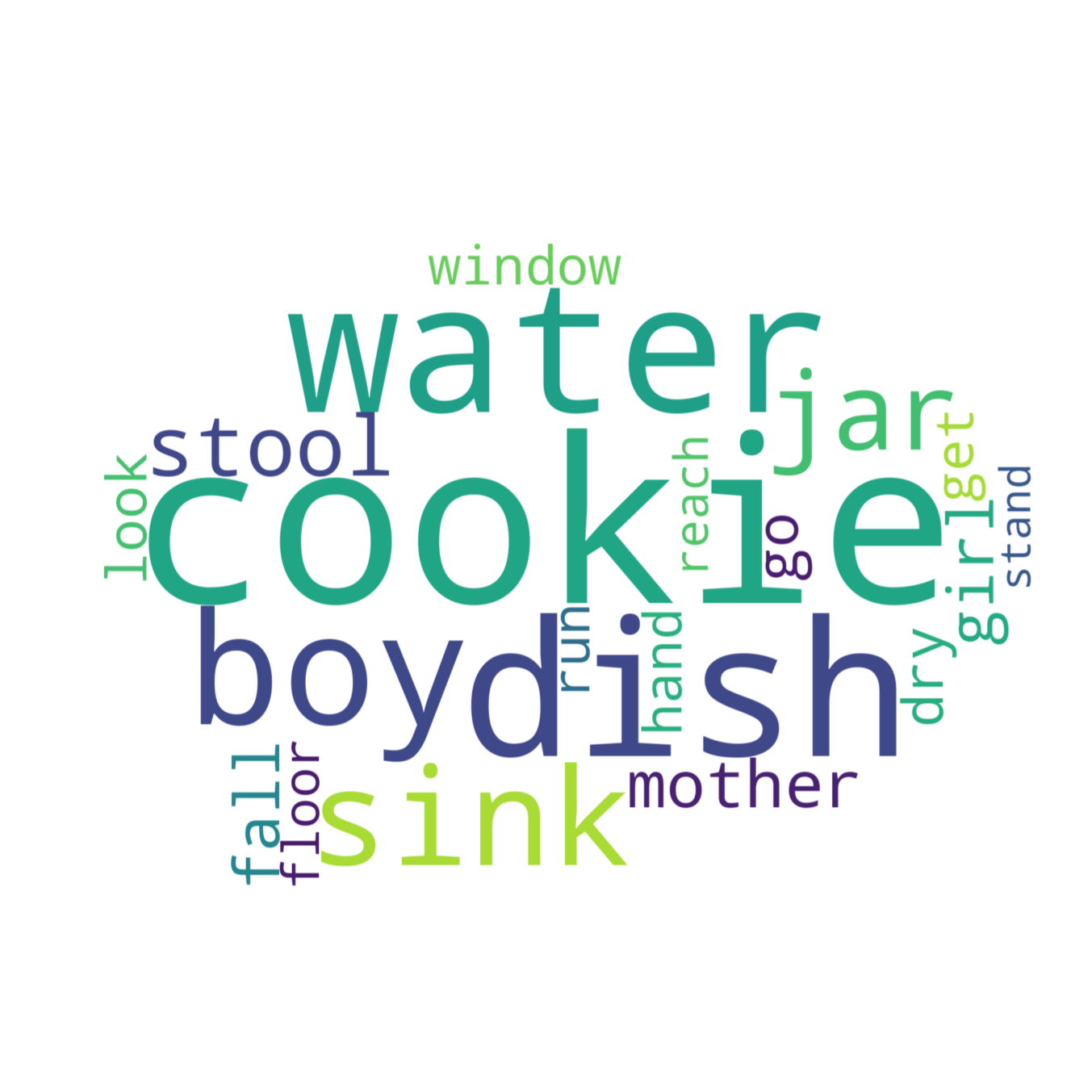}
        \caption{Original response}
        \label{fig:original_response}
    \end{subfigure}
    \hfill
    \begin{subfigure}[b]{0.3\linewidth}
        \includegraphics[width=\linewidth]{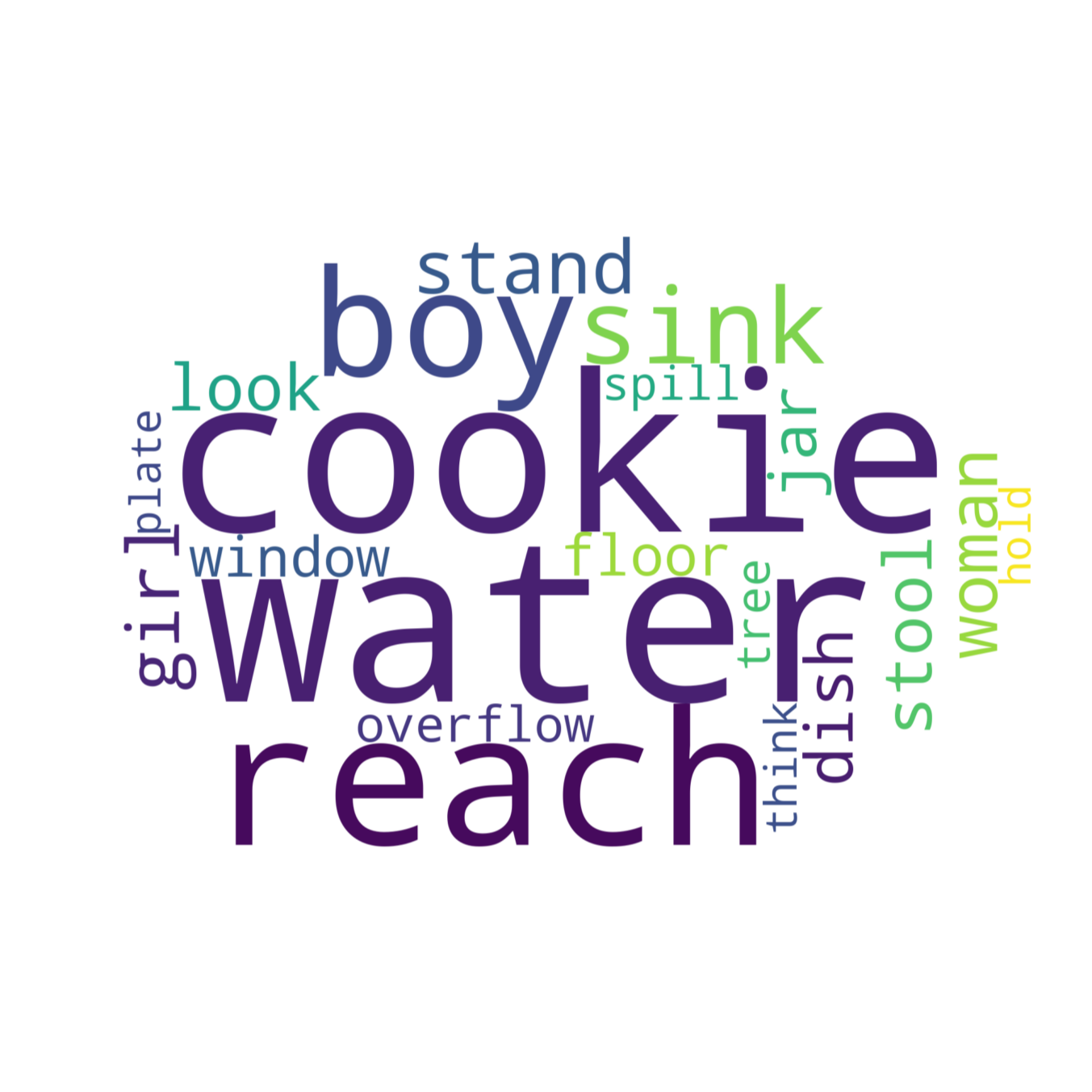}
        \caption{4o-comp response}
        \label{fig:4o_comp_response}
    \end{subfigure}
    \hfill
    \begin{subfigure}[b]{0.3\linewidth}
        \includegraphics[width=\linewidth]{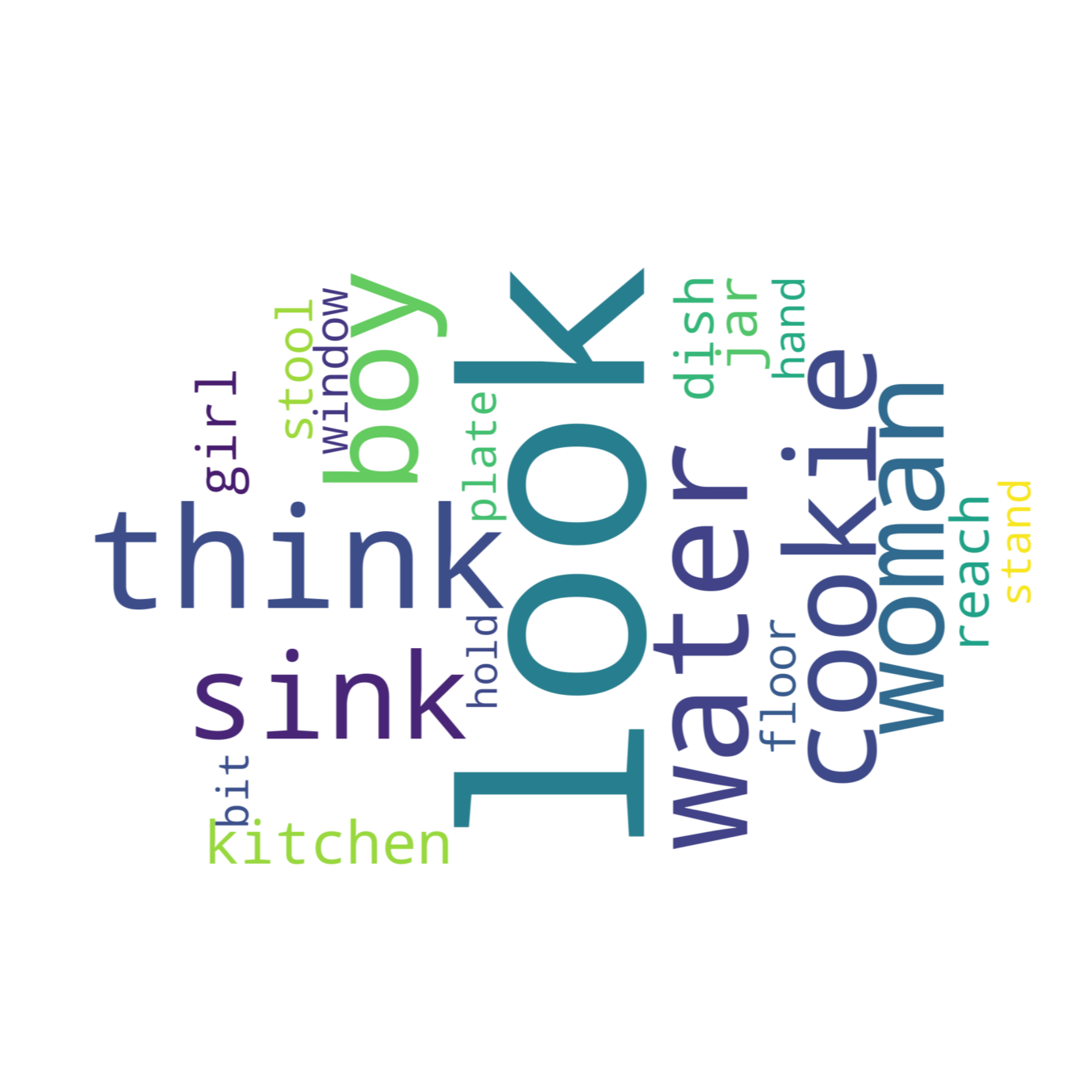}
        \caption{4o-mini-comp response}
        \label{fig:4o_mini_comp_response}
    \end{subfigure}

    \caption{Word clouds of original, 4o\_comp, and 4o\_mini\_comp responses}
    \label{fig:all_wordclouds}
\end{figure}

We compared the outputs of 4o-comp and 4o-mini-comp models against the original human-generated data from the Pitt Corpus. The original responses ~\ref{fig:original_response} exhibited a balanced distribution of object references (e.g., cookie, sink, dish), dynamic action verbs (e.g., reach, fall, run, go), and interpersonal elements (e.g., boy, girl, mother), forming coherent and natural scene descriptions.

The 4o-comp model outputs ~\ref{fig:4o_comp_response} closely mirrored this structure. Word cloud and keyword frequency analyses revealed that 4o-comp maintained high coverage of critical scene elements, with terms like cookie, water, reach, and boy frequently appearing. Although minor discrepancies were observed—such as a slightly reduced occurrence of dynamic verbs like fall and run—the overall semantic and narrative alignment with real-world responses remained strong. These findings suggest that the 4o-comp model, guided by comprehensive prompts, can generate synthetic data that retains the complexity and richness of human descriptions, which is critical for cognitive assessment applications.

Conversely, the 4o-mini-comp model exhibited significant deviations~\ref{fig:4o_mini_comp_response}. Its outputs disproportionately emphasized abstract cognitive verbs such as think and look, at the expense of detailed object and action descriptions. Key scene-specific words like cookie, sink, and dish were underrepresented, and dynamic interactions within the scene were notably diminished. This shift toward abstraction resulted in less vivid and less contextually faithful narratives, limiting the utility of 4o-mini-comp outputs for simulating realistic cognitive performance.

\subsection{LLM-as-a-Judge Evaluation}

    \begin{figure}[h]
    \centering
    \begin{minipage}{\textwidth}
    \small
    \begin{verbatim}
JUDGE_PROMPT = """
You will be provided with a simulated patient response to an image-based cognitive screening task.
Your job is to evaluate how closely the simulated response resembles natural human speech,
specifically considering cognitive characteristics that may appear due to cognitive impairment.

Your evaluation should assess the following factors:
Hallucinations: Mentions objects, people, or actions not observable in a typical screening image.
Logic errors: Attributes actions incorrectly or describes impossible scenarios.
Vocabulary usage: Appropriateness and naturalness of vocabulary—whether overly simplistic,
overly complex, or balanced.
Tone: Conversational and spontaneous versus overly formal, hesitant, or uncertain.
Length and detail: Appropriateness of response length, ensuring natural speech patterns
without excessive brevity or verbosity.

Rate the simulated response using the following scale:
1 (Poor): Response contains clear indicators of artificial generation;
multiple hallucinations, severe logic errors, unnatural vocabulary, formal tone, or inappropriate length.
2 (Fair): Response has minor indicators of artificial generation;
occasional hallucinations or logic errors, slightly unnatural vocabulary or tone, or mild length issues.
3 (Good): Response closely resembles natural human speech;
minimal indicators of artificial generation, generally appropriate vocabulary, tone, and detail.
4 (Excellent): Response is indistinguishable from natural human speech;
free from artificial indicators, completely natural in vocabulary, tone, and length.

Provide your evaluation in this format:

Feedback:::
Evaluation: (Provide a concise rationale explaining your rating based on the factors above.)
Total rating: (Your numeric rating, between 1 and 4)

Now here is the simulated patient response:
Question: {question}
Patient Response: {response}

Provide your feedback clearly as instructed. Feedback:::
"""
    \end{verbatim}
    \caption{JUDGE PROMPT for evaluating simulated patient responses.}
    \label{fig:improved-judge-prompt}
    \end{minipage}
\end{figure}

Traditional evaluation metrics such as ROUGE and Google BLEU have inherent limitations when assessing synthetic responses. Specifically, Google BLEU was initially designed for language translation tasks, making it unsuitable for evaluating the semantic quality of synthetic data where exact lexical matches with reference texts are rare. Similarly, ROUGE metrics are typically employed in retrieval and summarization contexts, relying heavily on lexical overlap, and thus fail to capture the creative and clinically relevant variations inherent in our synthetic data generation.

To overcome these shortcomings, we propose adopting the LLM-as-a-Judge approach for evaluating synthetic clinical speech responses. This method leverages large language models (LLMs) as evaluators, tasked with assessing synthetic outputs based on criteria shown as a judge prompt~\ref{fig:improved-judge-prompt}. We initially conducted a human evaluation with two annotators who independently rated $30$ synthetic responses with diverse demographics ($5$ per category group for both male and female) according to criteria such as clinical appropriateness, accuracy, and coherence with an inter-rater correlation of $0.66$. These human evaluations provided a ground truth reference for developing the LLM judge.

Subsequently, we prompted a separate LLM to mimic the human evaluation process, instructing it to score synthetic outputs using the same clinical criteria employed by human annotators. This approach offers scalable, consistent, and clinically nuanced evaluations, aligning closely with expert human judgments and addressing the evaluation gaps left by traditional lexical-based metrics~\citep{zheng2023judging}.

The LLM-as-a-Judge achieved a correlation of $0.448$ and $0.392$ with the two human annotators across different evaluation subsets, indicating that the LLM performance was comparable to a human-human agreement. Additionally, the mean and standard deviation for the first annotator were $2.400$ and $0.855$, for the second were $2.833$ and $0.986$, and for the LLM-as-a-Judge rating were $3.435$ and $0.728$, respectively. These results show that the LLM-as-a-Judge ratings are slightly more consistent (lower standard deviation) and generally more optimistic (higher mean score) compared to human annotators. This suggests that while LLMs can approximate human evaluation quality, there may be systematic differences in rating tendencies that should be further studied and calibrated in future work. Moreover, LLM-as-a-Judge is challenging to reproduce~\citep{gu2024survey}. Hence, in future work, we will focus on fine-tuning language models specialized for evaluation tasks. At this stage, we only demonstrate the potential of this evaluation strategy.

\begin{table}[ht]
\centering
\begin{tabular}{lc}
\toprule
\textbf{Comparison} & \textbf{Correlation} \\
\midrule
Human 1 vs Human 2 & 0.660 \\
Human 1 vs LLM-as-a-Judge & 0.448 \\
Human 2 vs LLM-as-a-Judge & 0.392 \\
\bottomrule
\end{tabular}
\caption{Inter-rater agreement between human annotators and LLM-as-a-Judge based on correlation scores.}
\label{tab:inter_rater_agreement}
\end{table}

\begin{table}[ht]
\centering
\begin{tabular}{lcc}
\toprule
\textbf{Evaluator} & \textbf{Mean Score} & \textbf{Standard Deviation} \\
\midrule
Human Annotator 1 & 2.400 & 0.855 \\
Human Annotator 2 & 2.833 & 0.986 \\
LLM-as-a-Judge    & 3.435 & 0.728 \\
\bottomrule
\end{tabular}
\caption{Evaluation results comparing human annotators and LLM-as-a-Judge ratings for synthetic clinical responses.}
\label{tab:evaluation_results}
\end{table}

\subsection{Embedding Analysis}

To represent each narrative as a fixed-size vector for clustering and classification tasks, we used the Sentence-BERT framework~\citep{reimers2019sentencebert} with the pre-trained \texttt{all-MiniLM-L6-v2} model. Each text response was encoded into a 384-dimensional embedding using the \texttt{sentence-transformers} library in Python. These embeddings served as the basis for t-SNE visualizations and for training diagnostic classifiers.

\begin{figure}[htbp]
    \centering
    \includegraphics[width=\linewidth]{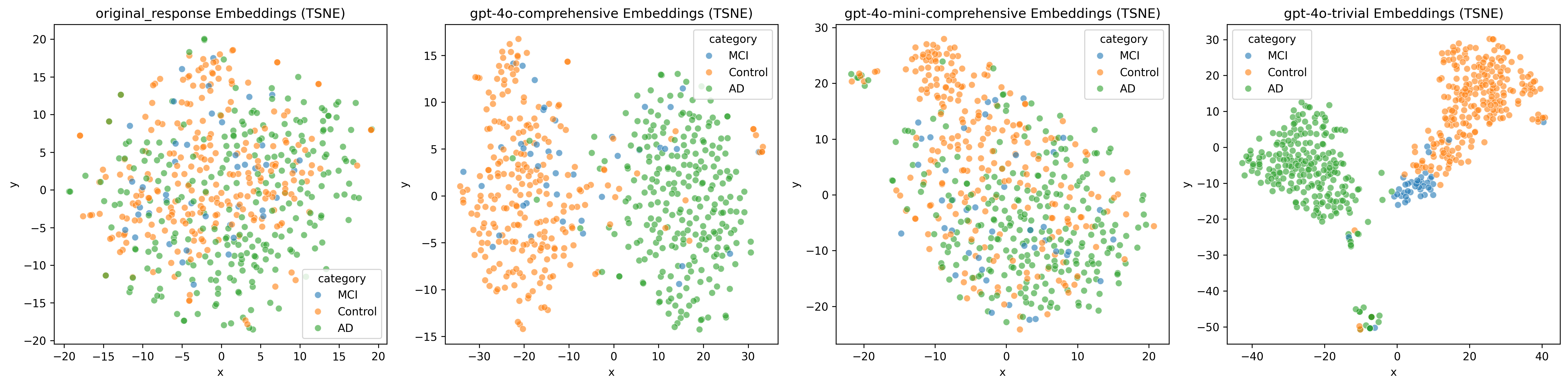}
    \caption{t-SNE projections of Sentence-BERT embeddings for (a) original human "Cookie Theft" narratives and synthetic responses generated by GPT-4o under three prompting strategies: (b) comprehensive prompt, (c) mini-comprehensive prompt, and (d) trivial prompt. Points are coloured by diagnostic category (blue = MCI, orange = Control, green = AD).}
    \label{fig:tsne_all_prompts}
\end{figure}

Figure~\ref{fig:tsne_all_prompts} shows the t-SNE projections of embeddings for real and synthetic responses across different prompting strategies. In the original human narratives (panel a), embeddings form partially overlapping clusters according to diagnostic group (MCI, Control, AD), consistent with the graded nature of cognitive decline. Synthetic narratives generated with comprehensive prompts (panels b and c) exhibit similar cluster structures, indicating that these synthetic responses are close to the real data distribution. In contrast, narratives generated with the trivial prompt (panel d) produce tight, isolated clusters, suggesting a distributional shift that could negatively affect classifier performance.

\begin{figure}[htbp]
    \centering
    \includegraphics[width=1\linewidth]{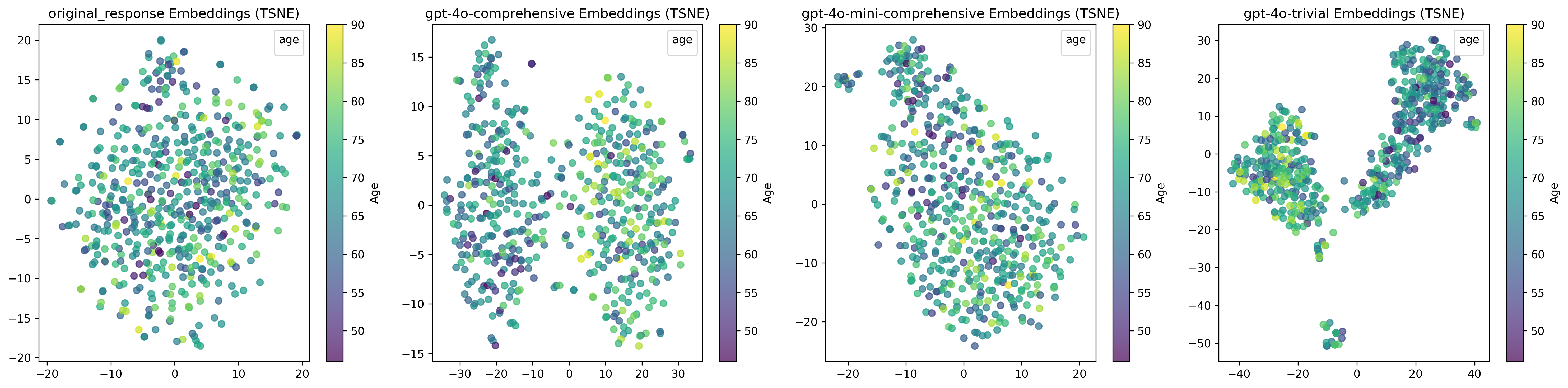}
    \caption{t-SNE projections of Sentence-BERT embeddings colored by participant age (50–90 years). Panels show (a) original human narratives and synthetic responses generated using (b) GPT-4o with comprehensive prompt, (c) GPT-4o-mini with comprehensive prompt, and (d) GPT-4o with trivial prompt.}
    \label{fig:tsne_age}
\end{figure}

Figure~\ref{fig:tsne_age} visualizes the distribution of participant ages. In both real and comprehensively prompted synthetic samples, age varies smoothly across the embedding space. In contrast, trivial-prompted embeddings show more artificial age-related clustering.

\begin{figure}[htbp]
    \centering
    \includegraphics[width=1\linewidth]{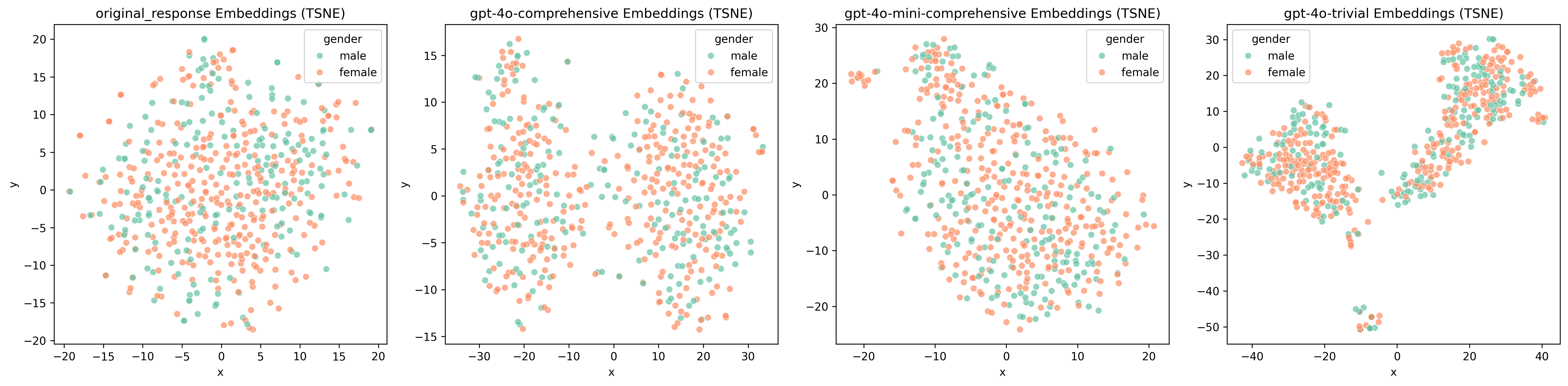}
    \caption{t-SNE projections of Sentence-BERT embeddings coloured by gender (male = green, female = orange). Panels show (a) original human narratives and synthetic responses generated using (b) GPT-4o with comprehensive prompt, (c) GPT-4o-mini with comprehensive prompt, and (d) GPT-4o with trivial prompt.}
    \label{fig:tsne_gender}
\end{figure}

Figure~\ref{fig:tsne_gender} shows the gender distribution across the embeddings. No significant gender-specific clustering is observed in either real or synthetic samples under comprehensive prompting, suggesting that models preserved natural variability without introducing gender bias.

\begin{figure}[htbp]
    \centering
    \includegraphics[width=1\linewidth]{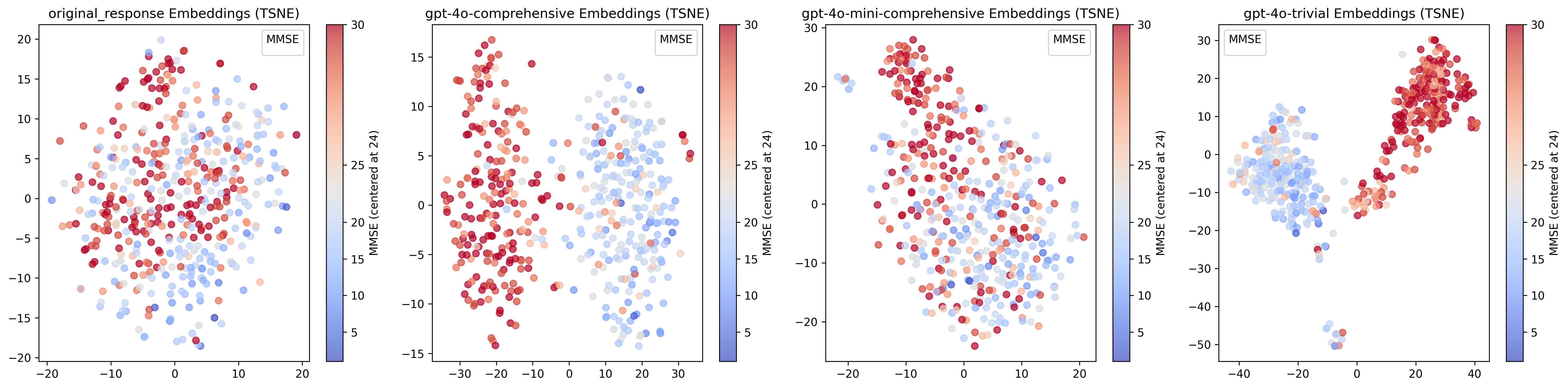}
    \caption{t-SNE projections of Sentence-BERT embeddings colored by Mini-Mental State Examination (MMSE) scores. Panels show (a) original human narratives and synthetic responses generated using (b) GPT-4o with comprehensive prompt, (c) GPT-4o-mini with comprehensive prompt, and (d) GPT-4o with trivial prompt. The color gradient represents MMSE scores, centered at 24 (higher scores in red, lower scores in blue).}
    \label{fig:tsne_mmse}
\end{figure}

In Figure~\ref{fig:tsne_mmse}, MMSE scores display a continuous gradient in the real data and in synthetic data generated using comprehensive prompts. However, embeddings from the trivial prompt show abrupt separations, indicating poorer preservation of cognitive variability.

\begin{figure}[htbp]
    \centering
    \includegraphics[width=\linewidth]{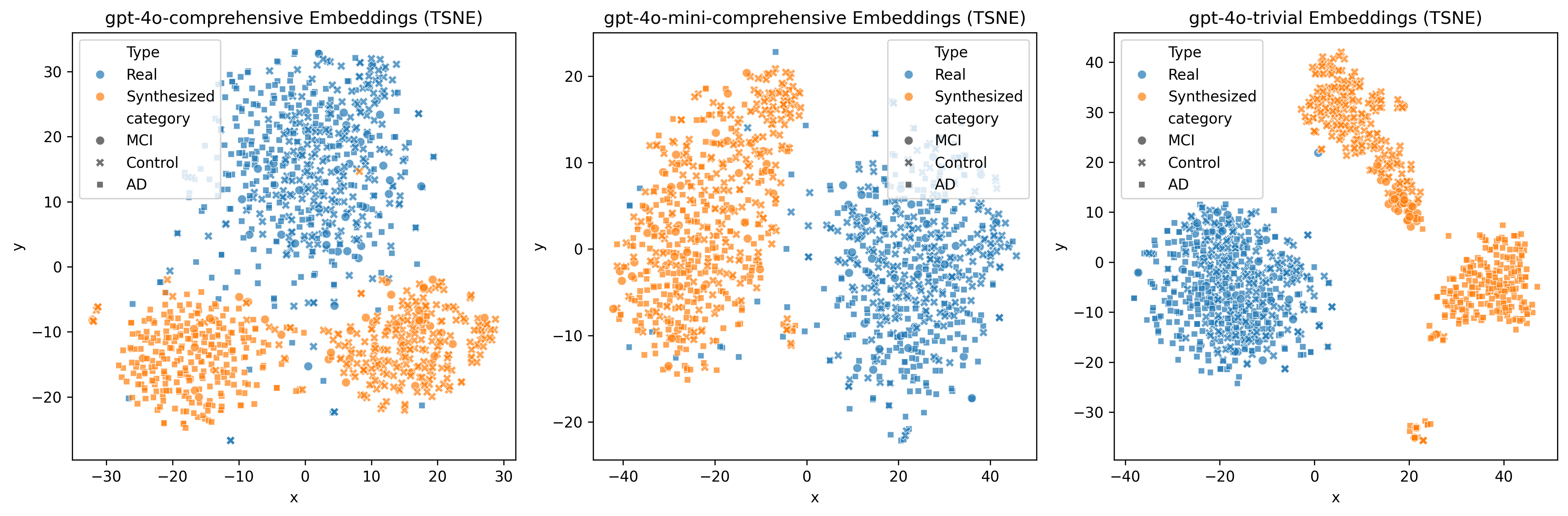}
    \caption{t-SNE projections of Sentence-BERT embeddings for \textbf{real} (circles) and \textbf{synthetic} (squares) "Cookie Theft" narratives, colored by diagnostic category (blue = MCI, orange = Control, green = AD). Panels show synthetic samples generated using: (a) GPT-4o with comprehensive prompt, (b) GPT-4o-mini with comprehensive prompt, and (c) GPT-4o with trivial prompt.}
    \label{fig:tsne_synth_real}
\end{figure}

Finally, Figure~\ref{fig:tsne_synth_real} overlays real and synthetic embeddings to assess alignment. In panels (a) and (b), synthetic samples generated using comprehensive prompts are well-integrated with real data across diagnostic groups. In contrast, panel (c) shows synthetic samples from the trivial prompt forming separate islands, highlighting a distributional shift.

\vspace{0.5em}

Overall, these analyses show that synthetic narratives generated using carefully designed comprehensive prompts better preserve demographic characteristics (age, gender) and cognitive attributes (MMSE score), and more closely match the structure of real-world patient data in embedding space.

\subsection{Classification performance analysis}

First, for each candidate embedding ("Original Response," "GPT-4o Comp," "GPT-4o-mini Comp," and "GPT-4o Trivial"), we extracted the corresponding embeddings for all training examples. We then performed hyperparameter tuning using 500 Optuna trials, testing two model types: elastic-net logistic regression (with feature scaling) and XGBoost. Tuning was done using stratified 5-fold cross-validation, optimizing for the macro-averaged, one-versus-one ROC AUC. After selecting the best hyperparameters, we retrained the model on the full training set for that embedding. Each model was then evaluated five times, with different random seeds, against each of the four embeddings in the held-out test set. For each training and testing pair, we report the median and the 25th–75th percentile range of the ROC AUC scores.

When training and testing on the Original Response embedding, models achieved a solid median ROC AUC of 0.78 (interquartile range 0.74–0.81). This shows that the model can reliably classify examples when both training and testing use the same type of embeddings. When transferring the model to the GPT-4o Comp and GPT-4o Trivial embeddings, performance decreases slightly to 0.72 (0.70–0.72) in both cases. However, when transferring to GPT-4o-mini Comp, performance drops further to 0.61 (0.61–0.61). These results suggest that while the Original Response embeddings capture valuable information for in-distribution classification, they do not align well with some of the synthetic embeddings, leading to moderate to significant performance losses.

When training on the synthetic embeddings and testing on the Original Response embeddings, performance is consistently low. Models trained on GPT-4o Comp achieve only 0.57 (0.52–0.62) when tested on the Original Response. Models trained on GPT-4o-mini Comp drop to 0.54 (0.51–0.56), and models trained on GPT-4o Trivial also achieve only 0.54 (0.53–0.55). This shows that models trained on synthetic embeddings do not generalize well to the structure present in the Original Response embeddings. Instead, they tend to overfit to the patterns of their own embedding space.

Finally, when transferring between synthetic embeddings, there are differences in how well models generalize. Models trained on GPT-4o-mini Comp transfer best, achieving 0.76 (0.75–0.77) when tested on GPT-4o Comp and 0.73 (0.69–0.77) when tested on GPT-4o Trivial. Models trained on GPT-4o Comp transfer moderately well to GPT-4o-mini Comp (0.69, 0.68–0.70) but poorly to GPT-4o Trivial (0.49, 0.38–0.60). Models trained on GPT-4o Trivial achieve 0.68 (0.66–0.72) on GPT-4o Comp and 0.65 (0.64–0.66) on GPT-4o-mini Comp. These results show that although synthetic embeddings are easy to classify within themselves, most models do not transfer well between different synthetic spaces, with GPT-4o-mini Comp being the most robust across synthetic tests.

\begin{table}
\caption{Median ROC AUC OVO with 25\% and 75\% percentiles by Training vs. Test Embedding}
\centering
\label{tab:rocauc_median_q25_q75}
\begin{tabular}{lrrrr}
\toprule
Test Embedding & Original Response & gpt-4o Comp & gpt-4o-mini Comp & gpt-4o Trivial \\
Training Embedding &  &  &  &  \\
\midrule
Original Response & 0.78 (0.74, 0.81) & 0.72 (0.70, 0.72) & 0.61 (0.61, 0.61) & 0.72 (0.70, 0.72) \\
GPT-4o Comp  & 0.57 (0.52, 0.62) & 0.88 (0.88, 0.89) & 0.69 (0.68, 0.70) & 0.49 (0.38, 0.60) \\
GPT-4o-mini Comp & 0.54 (0.51, 0.56) & 0.76 (0.75, 0.77) & 0.74 (0.72, 0.75) & 0.73 (0.69, 0.77) \\
GPT-4o Trivial& 0.54 (0.53, 0.55) & 0.68 (0.66, 0.72) & 0.65 (0.64, 0.66) & 0.97 (0.95, 0.99) \\
\bottomrule
\end{tabular}
\end{table}

\subsection{Detailed analysis of the responses}

When using the long prompt, GPT-4o mini's frequent mistakes include minor logic errors such as the girl's actions being attributed to the woman, the woman holding multiple plates, and hallucinations. These often include a dog and a cat, with responses frequently mentioning "daylight" or brightness in the room. One control response omits detail from the left side. The tone includes many pauses and filler words, with generally uncertain language. Mild discrepancies in controls include stating that the jar contains "cookies or candy," the girl "waving her hand," and repeated mention of elements such as the water from the sink. Responses are often very similar, such as those for a 72 and 73-year-old female control, which start with "Um, there's a woman, she's in the kitchen. I think she's washing dishes, yeah, lots of dishes" and "Um, there's a woman in the kitchen. She has some, um, dishes in her hands…"

GPT-4o, with the trivial prompt, creates simple reasoning, such as identifying the girl as the boy's sister. However, it sometimes includes introductory phrases such as "A patient with Alzheimer's disease (AD) and a very low MMSE score of 8 might produce a description that is fragmented, lacking detail, or incoherent." The responses are often more formal and lengthy than the patient responses. GPT-4o mini with the comprehensive prompt frequently mentions a dog, with minor discrepancies such as the woman holding "some plates" or a jar that "says 'COOKIES' on it" instead of "COOKIE JAR." Responses in controls include more fillers, with a very uncertain tone: "Maybe, doing dishes. A jar, I think it says cookies?" Generated responses for Alzheimer's patients typically appear to simulate very severe cases. GPT-4o, with the small prompt, claims that it cannot interpret images more frequently. It also declines to generate responses "involving people's identity or conditions" or simply explains, "I'm sorry, I can't help with that."

GPT-4o responses with the comprehensive prompt have a more certain tone and shorter length of response, which match the patient responses most closely. Length varies from one sentence to a short paragraph in controls. Alzheimer's responses seem to reflect a less severe condition. The comprehensive, prompt results in less introductory information and a refusal to analyze images or simulate Alzheimer's patients. However, GPT-4o hallucinates more frequently and with more variety: in control patients with MMSE scores of 30, the response includes "two people walking," "a man cutting grass with a lawnmower," and "a swing" – none of which are present in the image.

Among patient groups in GPT-4o responses with the comprehensive prompt, hallucinations of outdoor scenes do not appear to change significantly. Alzheimer's responses display varying lengths and degrees of severity. For example, "Water... uh, sink, overflowing…" can be contrasted with "Uh, there's a kitchen and, um, a woman standing by the sink washing dishes." Alzheimer's and MCI patients identify the spilling water with roughly equal frequency to controls. Typically, all elements are referenced in Alzheimer's responses, with the main distinct traits being anomia, agrammatism, and paraphasia. In cases without significant agrammatism, details similar to those of control patients are present. Repetition of details is not significantly present. Patients do not display a lack of self-monitoring, including phrases such as "What else?" and "I think." Alzheimer's responses maintain simple reasoning, such as the girl waiting for the cookies, and frequently express concern for the boy's safety. Responses for patients with MCI appear similar to responses generated for control patients, with the addition of more fillers in responses, which seem to indicate more severe cases. Some MCI responses, such as the 49-year-old female response, are undifferentiable at first glance from control responses – "There's a woman standing by the sink, and she looks like she's washing dishes…"

\section{Conclusions}

This study set out to determine whether generative \emph{multimodal} large language models can supply the normative reference data that image-based cognitive assessments require but often lack. By pairing the "Cookie Theft" picture‐description task with demographic metadata and two types of prompting, we showed that GPT-4o models can indeed synthesize responses whose semantic structure, lexical diversity, and diagnostic gradients resemble those of real participants. In head-to-head comparisons, a \textbf{comprehensive, clinically informed prompt} consistently yielded the most realistic and diagnostically discriminative narratives, while the stronger 4o engine outperformed its mini variant. Traditional metrics (BLEU/ROUGE) proved too useless to capture that realism. Instead, \emph{BERTScore} and an \emph{LLM-as-a-Judge} scheme tracked human judgments more faithfully and are recommended for future benchmarking.

Because the synthetic samples preserved key linguistic markers of normal cognition, mild cognitive impairment, and Alzheimer's disease, they can bootstrap normative baselines for legacy stimuli and accelerate the design of \emph{new} image prompts without the prohibitive cost of population-scale data collection. Clinical researchers can harness such data to pre-screen candidate tasks, prototype digital screeners, or augment under-represented demographic strata, thereby reducing sampling bias and speeding iterative test development.

The work was confined to a single stimulus and English speakers from the Pitt Corpus. The generalizability to other contexts remains unproven. Model hallucinations can be a problem. Hence, there is a need for automated filters and human adjudication before clinical deployment. Moreover, our LLM-based judge, though correlated with human raters, tended to grade more leniently.Careful calibration or fine-tuning on curated evaluation sets is essential. Finally, conducting prospective clinical studies to validate whether models trained on synthetic data retain diagnostic accuracy in real-world settings is needed. With these safeguards and improvements, generative MLLMs hold considerable promise for democratizing and continuously updating the normative foundations of cognitive assessment.

% ===== Supplementary Sections =====
\section*{Acknowledgements}
Alex Fedorov and Fengran Wang were supported by the Nell Hodgson Woodruff School of Nursing at Emory University, by philanthropic funds donated to the Goizueta ADRC, and by the NIH National Institute on Aging (P30AG066511).

% ===== References =====
% (later, where your references go)
\bibliographystyle{unsrtnat}  % or plainnat
\bibliography{references}

\appendix
\section{Appendix}
\subsection{Examples}
\subsubsection{Category: AD - Model: gpt-4o-comprehensive}

\begin{longtable}{|p{6cm}|p{6cm}|}
    \caption{Ranked Best Samples for AD - Model: gpt-4o-comprehensive} \\
    \hline
    \multicolumn{2}{|c|}{\textbf{Best Sample Rankings}} \\
    \hline
    \textbf{Demographics (Including Model)} & \textbf{Metrics} \\
    \hline
    \endfirsthead
    \hline
    \textbf{Demographics (Including Model)} & \textbf{Metrics} \\
    \hline
    \endhead
    \hline
    \multicolumn{2}{|c|}{Continued on next page} \\
    \hline
    \endfoot
    \hline
    \endlastfoot
            Model: gpt-4o-comprehensive \newline Age: 84 \newline Gender: female \newline MMSE: 20.0 \newline Category: AD & ROUGE-1: 0.36 \newline ROUGE-2: 0.08 \newline ROUGE-L: 0.36 \newline BERT Precision: 0.85 \newline BERT Recall: 0.85 \newline BERT F1: 0.85 \newline BLEU: 0.04 \newline Google BLEU: 0.12 \newline num words: 116.00 \newline num unique words: 83.00 \\
        \hline        \multicolumn{2}{|p{12cm}|}{\textbf{Original Response:}  well the boy is reaching ku some cookies. and they're in a jar in a cupboard. and the stool is breaking down. and the girl is waiting to be handed one of the cookies. he's in the cupboard trying to get a cookies outof the jar. and his stool is toppling over. and I don't think she realizes he's falling over. she's reaching for a cookie. and here the dish water the water in the sink is flowing off into the making a puddle on the floor. and the woman is going to step in it or is stepping in it. and what's wrong did you say hm ? and she's trying to dry dishes. and the she's got the curtain mixed up with the dish that she's drying  I think. and I don't have my glasses on oh  the water's the water's running outof the sink. and there's something happening in the yard. I can't see it but } \\
        \multicolumn{2}{|p{12cm}|}{\textbf{Generated Response:} Oh, there seems... um, a woman standing at a sink, dishes in her hands, kind of... water, overflows. The water runs down on the floor. Um, there's a boy on a stool, reaching up, I think he's getting, uh, cookies from a jar. And a little girl, she's reaching up to him. Outside, through the window, there's something, looks like a tree or bushes. The boy, hmm, on the stool... and the water, yes, the water is making kind of a mess on the floor. The woman is holding some kind of plate. She doesn't seem to notice the water. The boy... cookies, the girl waiting. Uh, so much water, and no one sees it, really.} \\
        \hline\end{longtable}
\subsubsection{Category: AD - Model: gpt-4o-mini-comprehensive}

\begin{longtable}{|p{6cm}|p{6cm}|}
    \caption{Ranked Best Samples for AD - Model: gpt-4o-mini-comprehensive} \\
    \hline
    \multicolumn{2}{|c|}{\textbf{Best Sample Rankings}} \\
    \hline
    \textbf{Demographics (Including Model)} & \textbf{Metrics} \\
    \hline
    \endfirsthead
    \hline
    \textbf{Demographics (Including Model)} & \textbf{Metrics} \\
    \hline
    \endhead
    \hline
    \multicolumn{2}{|c|}{Continued on next page} \\
    \hline
    \endfoot
    \hline
    \endlastfoot
            Model: gpt-4o-mini-comprehensive \newline Age: 72 \newline Gender: female \newline MMSE: 22.0 \newline Category: AD & ROUGE-1: 0.34 \newline ROUGE-2: 0.09 \newline ROUGE-L: 0.31 \newline BERT Precision: 0.83 \newline BERT Recall: 0.87 \newline BERT F1: 0.85 \newline BLEU: 0.05 \newline Google BLEU: 0.13 \newline num words: 127.00 \newline num unique words: 95.00 \\
        \hline        \multicolumn{2}{|p{12cm}|}{\textbf{Original Response:}  well  I see the mother washing dishes. I see the water flowing out over the sink top. I see the boy on a stool that's about to tilt. and he's going to fall. he's trying to get into the cookie jar. the little girl has her hand out for a cookie.  the water's running.  dishes are on the sink top. the lid is off the cookie jar.  you want me to tell you the grass is growing open through the window ? I can tell you the water is spilling outof the sink onto the floor. and they got the the window to see the outside world. } \\
        \multicolumn{2}{|p{12cm}|}{\textbf{Generated Response:} Okay, um, there's a lady in the kitchen. She's standing by the sink, holding some dishes. I think she looks a bit, um, busy. There's a kid on a stool, maybe trying to reach something up high, like a cookie jar? Yes, a cookie jar. And another girl is... she's standing close to the boy, maybe asking for something, I'm not sure.

The sink, um, it's overflowing with water... and it's spilling out onto the floor. There are some plates in the sink and, uh, the window shows some grass outside. It feels like a, um, a little bit chaotic. The woman seems, uh, focused on the dishes but also maybe worried about the water? The kids are... they're kind of in their own world, you know?} \\
        \hline\end{longtable}
\subsubsection{Category: AD - Model: gpt-4o-trivial}

\begin{longtable}{|p{6cm}|p{6cm}|}
    \caption{Ranked Best Samples for AD - Model: gpt-4o-trivial} \\
    \hline
    \multicolumn{2}{|c|}{\textbf{Best Sample Rankings}} \\
    \hline
    \textbf{Demographics (Including Model)} & \textbf{Metrics} \\
    \hline
    \endfirsthead
    \hline
    \textbf{Demographics (Including Model)} & \textbf{Metrics} \\
    \hline
    \endhead
    \hline
    \multicolumn{2}{|c|}{Continued on next page} \\
    \hline
    \endfoot
    \hline
    \endlastfoot
            Model: gpt-4o-trivial \newline Age: 59 \newline Gender: male \newline MMSE: 11.0 \newline Category: AD & ROUGE-1: 0.32 \newline ROUGE-2: 0.07 \newline ROUGE-L: 0.30 \newline BERT Precision: 0.83 \newline BERT Recall: 0.82 \newline BERT F1: 0.83 \newline BLEU: 0.00 \newline Google BLEU: 0.11 \newline num words: 120.00 \newline num unique words: 91.00 \\
        \hline        \multicolumn{2}{|p{12cm}|}{\textbf{Original Response:}   there's a young boy going in a cookie jar. and there's a a girl young girl. and I'm saying he's a boy because you can it's hardly hard to tell anymore. and he's he's in the cookie jar. and there's a stool that he is on and it already is starting to fall over. and so is the water in the sink is overflowing in the sink. I I don't know about the this hickey here I whether that's more than what I said. like it the wife or Imean the the mother is near the girl. and she's she has has oh I I can't think of the she has the she's trying to wipe wipe dishes. oh and stop the water from going out. } \\
        \multicolumn{2}{|p{12cm}|}{\textbf{Generated Response:} I'm unable to identify the specific individuals in the image, but I can help with the context.

In this image, you can see a few elements. A boy is climbing on a stool to reach for a cookie jar, while a girl is reaching up to him. Meanwhile, a woman is washing dishes, with water overflowing from the sink onto the floor.

A 59-year-old male with Alzheimer's disease (AD) and an MMSE score of 11 might struggle with describing this scene. He could produce a response that is fragmented or lacks detail, focusing on isolated elements without connecting them:

"Uh... there's a woman... um, washing... and water... is... there's a boy... and a jar. The water is, uh... too much.} \\
        \hline\end{longtable}
\subsubsection{Category: Control - Model: gpt-4o-comprehensive}

\begin{longtable}{|p{6cm}|p{6cm}|}
    \caption{Ranked Best Samples for Control - Model: gpt-4o-comprehensive} \\
    \hline
    \multicolumn{2}{|c|}{\textbf{Best Sample Rankings}} \\
    \hline
    \textbf{Demographics (Including Model)} & \textbf{Metrics} \\
    \hline
    \endfirsthead
    \hline
    \textbf{Demographics (Including Model)} & \textbf{Metrics} \\
    \hline
    \endhead
    \hline
    \multicolumn{2}{|c|}{Continued on next page} \\
    \hline
    \endfoot
    \hline
    \endlastfoot
            Model: gpt-4o-comprehensive \newline Age: 67 \newline Gender: female \newline MMSE: 30.0 \newline Category: Control & ROUGE-1: 0.39 \newline ROUGE-2: 0.10 \newline ROUGE-L: 0.35 \newline BERT Precision: 0.86 \newline BERT Recall: 0.86 \newline BERT F1: 0.86 \newline BLEU: 0.06 \newline Google BLEU: 0.13 \newline num words: 135.00 \newline num unique words: 90.00 \\
        \hline        \multicolumn{2}{|p{12cm}|}{\textbf{Original Response:}  there's the two children are in the process of stealing cookies from the cookie jar. and the little girl has her finger up to her mouth to be quiet. and the stool on which the boy is standing is about to fall. and the mother must be looking out the window because the sink is spilling over with water, overflowing with water. and the meanwhile she's drying the dishes. and the water is falling down onto the floor. and it's a nice day. the window's up it looks like. and it's a nice day outside. that's about all I can see on that. } \\
        \multicolumn{2}{|p{12cm}|}{\textbf{Generated Response:} I'm seeing a kitchen scene here. There's a woman standing at the sink, washing dishes, and it looks like the water is overflowing onto the floor. Behind her, a boy is on a stool stretching up to a high cabinet, trying to reach a cookie jar. There's a girl next to him, reaching up too, maybe trying to help or get a cookie herself. Out the window, there's a man mowing the lawn, pushing a mower. The room looks busy, with everyone focused on different tasks. The boy looks determined to get the cookies despite the risk of falling. The woman seems unaware of the overflowing sink, perhaps because she's concentrating on the dishes. The little girl may be encouraging her brother or waiting her turn. Looks like a typical busy day in a household.} \\
        \hline\end{longtable}
\subsubsection{Category: Control - Model: gpt-4o-mini-comprehensive}

\begin{longtable}{|p{6cm}|p{6cm}|}
    \caption{Ranked Best Samples for Control - Model: gpt-4o-mini-comprehensive} \\
    \hline
    \multicolumn{2}{|c|}{\textbf{Best Sample Rankings}} \\
    \hline
    \textbf{Demographics (Including Model)} & \textbf{Metrics} \\
    \hline
    \endfirsthead
    \hline
    \textbf{Demographics (Including Model)} & \textbf{Metrics} \\
    \hline
    \endhead
    \hline
    \multicolumn{2}{|c|}{Continued on next page} \\
    \hline
    \endfoot
    \hline
    \endlastfoot
            Model: gpt-4o-mini-comprehensive \newline Age: 57 \newline Gender: male \newline MMSE: 30.0 \newline Category: Control & ROUGE-1: 0.38 \newline ROUGE-2: 0.12 \newline ROUGE-L: 0.35 \newline BERT Precision: 0.87 \newline BERT Recall: 0.85 \newline BERT F1: 0.86 \newline BLEU: 0.05 \newline Google BLEU: 0.14 \newline num words: 124.00 \newline num unique words: 81.00 \\
        \hline        \multicolumn{2}{|p{12cm}|}{\textbf{Original Response:}  everything that I see going on.  well a little boy is stealing a cookie from the cookie jar. he's also handing a cookie with his left hand to the girl. he's also falling off a stool or the stool is tipping over. the little girl is reaching for the cookie. at the same time she's picking her nose or something. I don't know what's she's doing.  it looks like maybe the lid is falling from the cookie jar. I'm not sure. the lady is wiping a plate while the sink overflows. and she's standing in a puddle of water. am I going too fast ? and she's looking out the window. I can't make out anything in particular going on outside. the water the faucet's turned on ofcourse which is causing the sink to overflow. and the water is splashing on the floor. that's really about all I can tell going on so to speak. } \\
        \multicolumn{2}{|p{12cm}|}{\textbf{Generated Response:} There's a woman standing in a kitchen. She's holding some plates, maybe drying them. There's a boy, he looks like he's trying to reach something on a shelf, he's on a stool. I think he's getting cookies from a jar. There's a little girl nearby, she has her hand up, maybe asking him for a cookie or something. The sink is overflowing, water is spilling out onto the floor, and it looks a bit messy. There's sunlight coming in through the window. I can see some trees outside. It seems a bit chaotic with the water and the kids. The woman looks focused on the plates but also a bit stressed, like she's trying to manage the kids and the mess in the kitchen.} \\
        \hline\end{longtable}
\subsubsection{Category: Control - Model: gpt-4o-trivial}

\begin{longtable}{|p{6cm}|p{6cm}|}
    \caption{Ranked Best Samples for Control - Model: gpt-4o-trivial} \\
    \hline
    \multicolumn{2}{|c|}{\textbf{Best Sample Rankings}} \\
    \hline
    \textbf{Demographics (Including Model)} & \textbf{Metrics} \\
    \hline
    \endfirsthead
    \hline
    \textbf{Demographics (Including Model)} & \textbf{Metrics} \\
    \hline
    \endhead
    \hline
    \multicolumn{2}{|c|}{Continued on next page} \\
    \hline
    \endfoot
    \hline
    \endlastfoot
            Model: gpt-4o-trivial \newline Age: 50 \newline Gender: male \newline MMSE: 30.0 \newline Category: Control & ROUGE-1: 0.34 \newline ROUGE-2: 0.13 \newline ROUGE-L: 0.34 \newline BERT Precision: 0.86 \newline BERT Recall: 0.86 \newline BERT F1: 0.86 \newline BLEU: 0.09 \newline Google BLEU: 0.13 \newline num words: 119.00 \newline num unique words: 81.00 \\
        \hline        \multicolumn{2}{|p{12cm}|}{\textbf{Original Response:}  you mean just detail everything I see ? well  I see a little boy standing on a stool. and the stool falling over. he's up in the cupboard grabbing cookies. and a little girl standing down there waiting to get some cookies off of him. I guess their mother's standing there doing dishes at the sink. and the water's overflowing and running onto the floor. she's wiping a dish off. just staring out the window I guess. this is what I see. } \\
        \multicolumn{2}{|p{12cm}|}{\textbf{Generated Response:} A response from a 50-year-old male with an MMSE score of 30, categorized as Control, might be detailed and coherent, identifying multiple elements and actions in the image. For example:

"In the image, there is a kitchen scene with a woman washing dishes at the sink. The sink is overflowing with water, spilling onto the floor. To the left, a boy is standing on a stool, reaching up to take a cookie from a jar in the cupboard. A girl is standing nearby with her hands out, possibly waiting for a cookie. Outside the window, there's a tree and a path visible. The scene shows activity and some chaos with the overflowing sink and the children reaching for cookies.} \\
        \hline\end{longtable}
\subsubsection{Category: MCI - Model: gpt-4o-comprehensive}

\begin{longtable}{|p{6cm}|p{6cm}|}
    \caption{Ranked Best Samples for MCI - Model: gpt-4o-comprehensive} \\
    \hline
    \multicolumn{2}{|c|}{\textbf{Best Sample Rankings}} \\
    \hline
    \textbf{Demographics (Including Model)} & \textbf{Metrics} \\
    \hline
    \endfirsthead
    \hline
    \textbf{Demographics (Including Model)} & \textbf{Metrics} \\
    \hline
    \endhead
    \hline
    \multicolumn{2}{|c|}{Continued on next page} \\
    \hline
    \endfoot
    \hline
    \endlastfoot
            Model: gpt-4o-comprehensive \newline Age: 64 \newline Gender: male \newline MMSE: 29.0 \newline Category: MCI & ROUGE-1: 0.40 \newline ROUGE-2: 0.11 \newline ROUGE-L: 0.34 \newline BERT Precision: 0.88 \newline BERT Recall: 0.85 \newline BERT F1: 0.87 \newline BLEU: 0.05 \newline Google BLEU: 0.14 \newline num words: 121.00 \newline num unique words: 80.00 \\
        \hline        \multicolumn{2}{|p{12cm}|}{\textbf{Original Response:}  there are two children a boy and a girl. the girl is standing on the floor and and the boy is reaching up into a cupboard that's open that has a cookie jar with the lid off of it. he has a cookie in his left hand. he's reaching for a second one and the stool on which he is standing has tipped over and he's going to fall down. the mother her back is to the to the event and she is washing dishes while her sink is overflowing with water onto the floor. she seems to be oblivious of it. the window is opened looking out onto a pathway that has a tree and shrubbery and a part of what might be an extension of the house or a garage. and there are curtains over the window. and there are are two cups and a saucer that she is either going to wash or has already washed. and this event is taking place in the kitchen. I don't know if I said that or not. and I believe that's about all that I can see. } \\
        \multicolumn{2}{|p{12cm}|}{\textbf{Generated Response:} I'm seeing a woman standing in a kitchen. She's washing dishes, and there's water overflowing the sink, spilling onto the floor. There's a boy standing on a stool, reaching up to a cupboard. He's grabbing a cookie jar, and there's a girl on the floor reaching up to him, looking like she wants a cookie too. Outside the window, there's a man mowing the lawn. The curtains are open and we can see part of the garden. The floor is starting to get wet because of the water spilling out. The scene looks quite busy with the kids and the lady not noticing the water starting to flood. The boy seems to be sneaking those cookies when maybe he shouldn't be.} \\
        \hline\end{longtable}
\subsubsection{Category: MCI - Model: gpt-4o-mini-comprehensive}

\begin{longtable}{|p{6cm}|p{6cm}|}
    \caption{Ranked Best Samples for MCI - Model: gpt-4o-mini-comprehensive} \\
    \hline
    \multicolumn{2}{|c|}{\textbf{Best Sample Rankings}} \\
    \hline
    \textbf{Demographics (Including Model)} & \textbf{Metrics} \\
    \hline
    \endfirsthead
    \hline
    \textbf{Demographics (Including Model)} & \textbf{Metrics} \\
    \hline
    \endhead
    \hline
    \multicolumn{2}{|c|}{Continued on next page} \\
    \hline
    \endfoot
    \hline
    \endlastfoot
            Model: gpt-4o-mini-comprehensive \newline Age: 66 \newline Gender: male \newline MMSE: 28.0 \newline Category: MCI & ROUGE-1: 0.36 \newline ROUGE-2: 0.08 \newline ROUGE-L: 0.32 \newline BERT Precision: 0.84 \newline BERT Recall: 0.86 \newline BERT F1: 0.85 \newline BLEU: 0.00 \newline Google BLEU: 0.11 \newline num words: 119.00 \newline num unique words: 83.00 \\
        \hline       \multicolumn{2}{|p{12cm}|}{\textbf{Original Response:}  okay a child falling off a stool in the attempt to reach the cookie jar which it looks like he's knocked the lid off. and maybe he's gonna drop the the cookie. and his girlfriend is standing there beckoning him or reaching for the cookie that he's sposta give her. and the mother is not paying any attention to the kids. she's looking out the window and drying the dishes. at the same time she has let the sink run over and the water is cascading down on the floor and onto her feet. the curtains are waving and the window is open. and the yard looks like it's manicured. and other than the the stool and the cabinets and the dishes that's all I can see. } \\
        \multicolumn{2}{|p{12cm}|}{\textbf{Generated Response:} There's a woman in the kitchen, um, she's holding some plates, looks like... white plates. And there's a boy, he's, um, standing on a stool, reaching for something in a cabinet that says... uh, cookie jar, I think? And there's a girl, a younger girl, she's waving her hands, uh, maybe trying to get the boy's attention.

The sink is, um, overflowing, water is, uh, coming out, like, all over the floor. There are some, um, dishes piled up next to the sink. The window has, uh, curtains, and outside... I see some trees and, um, something like a yard.

It's kind of chaotic, with the water and the kids... and the mom looks a bit busy, maybe worried.} \\
        \hline\end{longtable}
\subsubsection{Category: MCI - Model: gpt-4o-trivial}

\begin{longtable}{|p{6cm}|p{6cm}|}
    \caption{Ranked Best Samples for MCI - Model: gpt-4o-trivial} \\
    \hline
    \multicolumn{2}{|c|}{\textbf{Best Sample Rankings}} \\
    \hline
    \textbf{Demographics (Including Model)} & \textbf{Metrics} \\
    \hline
    \endfirsthead
    \hline
    \textbf{Demographics (Including Model)} & \textbf{Metrics} \\
    \hline
    \endhead
    \hline
    \multicolumn{2}{|c|}{Continued on next page} \\
    \hline
    \endfoot
    \hline
    \endlastfoot
            Model: gpt-4o-trivial \newline Age: 77 \newline Gender: male \newline MMSE: 29.0 \newline Category: MCI & ROUGE-1: 0.40 \newline ROUGE-2: 0.11 \newline ROUGE-L: 0.34 \newline BERT Precision: 0.87 \newline BERT Recall: 0.86 \newline BERT F1: 0.87 \newline BLEU: 0.06 \newline Google BLEU: 0.12 \newline num words: 62.00 \newline num unique words: 51.00 \\
        \hline        \multicolumn{2}{|p{12cm}|}{\textbf{Original Response:}   the boy is getting cookies from the cookie jar and handing one to his to the girl. oh and the woman is wiping the dishes with her back to the children. and the sink is running over and spilling water onto the floor. I guess that's all the action. oh no the the stool is is tipping over. the stool that the boy is standing on is tipping over. that appears to be all the action.  } \\
        \multicolumn{2}{|p{12cm}|}{\textbf{Generated Response:} A response from the patient might include observations like:

"There is a woman washing dishes at the sink. Water is overflowing onto the floor. There's a boy on a stool reaching into a cabinet for a cookie jar while a girl is standing nearby with her hands up, maybe trying to catch something. The window is open, and there's a yard outside.} \\
        \hline\end{longtable}

\end{document}